\documentclass[10pt,twocolumn,letterpaper=true]{article}

\usepackage{cvpr}
\usepackage{times}
\usepackage{epsfig}
\usepackage{graphicx}
\usepackage{amsmath}
\usepackage{amssymb}

\usepackage[pagebackref=true,breaklinks=true,colorlinks,bookmarks=false,letterpaper=true]{hyperref}

\cvprfinalcopy 

\def\cvprPaperID{0027} 

\begin{document}

\title{Automated Tackle Injury Risk Assessment in Contact-Based Sports - A Rugby Union Example}

\author{Zubair Martin\\
African Robotics Unit, University of Cape Town\\
{\tt\small MRTZUB001@myuct.ac.za}
\and
Sharief Hendricks\\
Exercise Science \& Sports Medicine Division,\\ 
University of Cape Town\\
\and
Amir Patel\\
African Robotics Unit, University of Cape Town\\
}

\maketitle

\begin{abstract}
Video analysis in tackle-collision based sports is highly subjective and exposed to bias, which is inherent in human observation, especially under time constraints. This limitation of match analysis in tackle-collision based sports can be seen as an opportunity for computer vision applications. Objectively tracking, detecting and recognising an athlete’s movements and actions during match play from a distance using video, along with our improved understanding of injury aetiology and skill execution will enhance our understanding how injury occurs, assist match day injury management, reduce referee subjectivity. In this paper, we present a system of objectively evaluating in-game tackle risk in rugby union matches. First, a ball detection model is trained using the You Only Look Once (YOLO) framework, these detections are then tracked by a Kalman Filter (KF). Following this, a separate YOLO model is used to detect persons/players within a tackle segment and then the ball-carrier and tackler are identified. Subsequently, we utilize OpenPose to determine the pose of ball-carrier and tackle, the relative pose of these is then used to evaluate the risk of the tackle. We tested the system on a diverse collection of rugby tackles and achieved an evaluation accuracy of 62.50\%. These results will enable referees in tackle-contact based sports to make more subjective decisions, ultimately making these sports safer.
\end{abstract}
\section{Introduction}
The tackle is a physically and technically dynamic contest between two opposing players for territory and ball possession. The tackle event is common in all tackle-contact based sports, such as Rugby Union, Rugby Sevens, Rugby League and American Football. During a tackle, the defending player(s), known as the tackler, attempts to impede the attacker’s (the ball-carrier) progression towards the goal-line and regain possession of the ball \cite{hendricks2020consensus, quarrie2008tackle, fuller2010injury}. This action is called tackling  and is the main form of defence in tackle-collision based sports. As a result, it is the most frequently occurring contact event in tackle-collision based sports and associated with player and team performance \cite{hendricks2021tackle, burger2020lay}.  For example, in Rugby League, the tackle event can occur up to 700 times in an 80-minute match \cite{king2010video}. At the same time, the frequency of occurrence and technical-physical nature of the tackle places both the ball-carrier and tackler at high-risk of injury. In the Rugby Union, tackle injuries have the highest injury incidence \cite{williams2013meta} - which may cause the greatest number of days lost (severity) \cite{brooks2005epidemiology} and carry a high injury burden (injury incidence rate × mean days absent per injury)\cite{west2020trends}. 

\begin{figure}[t]
\fbox{\includegraphics[width=1\linewidth]{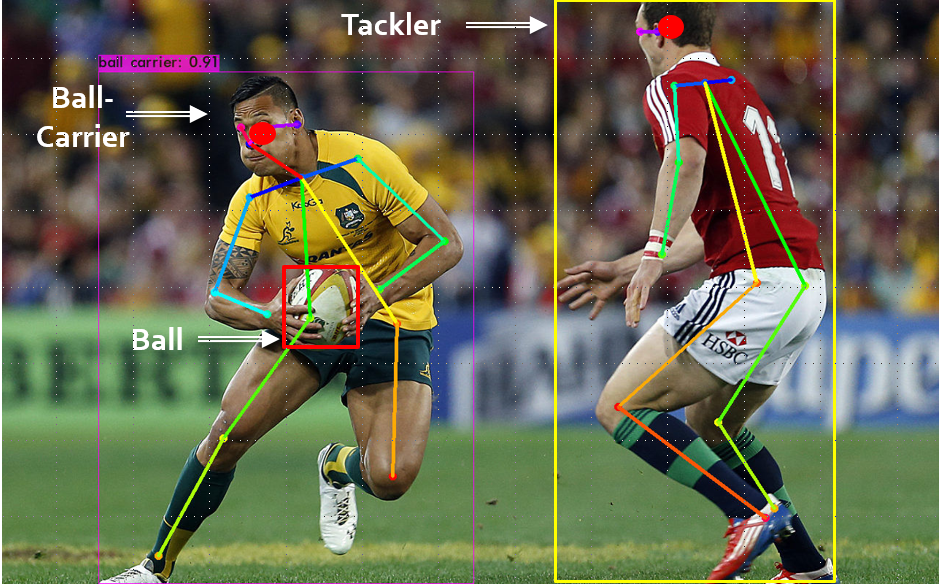}}\\
   \caption{Illustration of the tackle frame
   - highlighting the ball, tackler and ball-carrier detection. The figure further displays the key point estimates of both players (outline of skeleton) and estimate of the head-centre of both players (red dot). }
\label{fig:tackle_frame}
\end{figure}

Head injuries during the tackle specifically, is a major concern for these tackle collision-based sports. The overall incidence (injuries/1000 playing hours) of concussion in the rugby union, rugby sevens and rugby league are 4.7\cite{gardner2014systematic} , 8.3 \cite{fuller2015epidemiology} and 8.0-17.5 \cite{gardner2015systematic}, respectively. Preventing these head injuries will require a multi-faceted approach – understanding injury mechanisms and risk factors, driving player, coach, referee and stakeholder education, advancing how we train the tackle, and modifying the laws of the game. A key risk factor for tackle head injuries is ball-carrier and tackler technique (13-17) . Specifically, tacklers are not moving from an upright to low body position, thereby increasing the likelihood of head to head contact \cite{tucker2017risk}. 

Recently, to nudge tacklers to modify their tackling technique and reduce their tackling height, World Rugby (WR), rugby union’s governing body, have implemented stricter rulings around high tackles and trialled a law change to lower the height of a legal tackle, from above the line of the shoulders to above the line of the armpit \cite{raftery2020getting, stokes2021does}. Implementing these law change injury prevention strategies is highly reliant on referee decision-making, and in 2018, the second season of stricter high tackle sanctioning, WR noted intra-competition and inter-competition inconsistencies, especially when referees issued yellow and red cards. These inconsistencies exposed WR to public criticism, which can diminish the potential effectiveness of an injury prevention initiative. In response and to improve consistency and fairness, WR introduced the high tackle decision making framework – a flow chart to assist match day officials in sanctioning high-tackles \cite{raftery2020getting}. The match day referee and assistant referees can apply this decision making framework in real-time or when reviewing video replays (in slow motion and from different angles). Referee decisions however, are influenced by a number of factors such as decision-making time, home advantage and crowd \cite{spitz2021video}. A system that can objectively classify tackle head injury risk may assist referees with their decision-making. Spectators, players and coaches can also then view the decision-making process of the referee.

In the world of rugby, the final referee rulings are based on whether a tackle was legal or illegal. From a coaching technical perspective, not all high-risk tackles are illegal. However, they do influence the referee’s decision as they are dangerous and prone to injury. According to an analyst with expertise and experience in rugby video analysis, a high-risk tackle is defined as a tackle in which the heads of the tackler and ball-carrier are in proximity \cite{hendricks_2020}. These are identified by the ball-carrier and tackler having the same body position when contact is made (upright or low). On the other hand, a low-risk tackle is defined as one whereby the body positions differ. The risk is referred to any player involved (overall) and not specific to the ball-carrier or tackler \cite{hendricks_2020}.

Video footage plays an important role in the world of rugby. Throughout the match, team doctors compile a video analysis of concussion incidents - with the goal of preventing future concussive injuries \cite{10}. Furthermore, television match officials (TMOs) make use of video footage to assist the referee in making a fair decision \cite{11}. Therefore, the aim of this paper is to explore computer vision solutions to address an alternative strategy of concussion prevention (using in-game live video footage). The objective is to identify a high and low-risk tackle which in future may aid coaches in improved training tackle techniques and referees in generating an objective decision. 

Video analysis plays a vital role in understanding injury mechanisms and risk factors, and assisting medical protocols (e.g., head injury assessment) \cite{fuller2017accuracy, gardner2018use} and referee decision making (e.g., television match official reviewing dangerous play). However, video analysis for tackle-collision based sports still remains subjective. In this paper, we present a method of automatic tackle risk assessment for rugby union.  This system automatically detects the ball, ball-carrier and tackler and uses their relative pose to predict tackle risk as depicted in Figure \ref{fig:tackle_frame}.

\section{Related Works}
There is seen to be a great opportunity in computer vision applications of capturing statistical data in sports. The use of capturing data at a distance - without any contact, allows for a non-invasive strategy whereby the actions of the players are not influenced \cite{12}. With the advantage of tracking, detection and recognition - the understanding of injuries can be improved \cite{12}.

A recent approach in detecting tackle risk (specifically concussions) embeds sensors within the player kits (accelerometers)  \cite{14}. The main downfall of this approach is that acceleration alone does not provide enough information to identify tackle risk \cite{14}. Implementing additional sensors can become invasive to the natural behaviour of players and therefore would influence the collected data. Other methods such as marker based motion capture (Vicon) require multiple marker key points which may result in high cost and limited to a control test environment (as shown in \cite{15}). 

One study investigated a non-invasive method (Organic Motion markerless motion tracking) in determining the kinematic measurements of players who are at high-risk of concussion \cite{14}. The experiments conducted focused on testing the cognitive behaviour of four players in a given obstacle course \cite{14}. Video capture of the experiments were collected using 16 stationary cameras by tracking key player points. To our knowledge, there is currently no previous study which utilized computer vision for tackle risk evaluation (high or low-risk) from in-game (live) footage.

\begin{figure*}
\fbox{\includegraphics[width=1\linewidth]{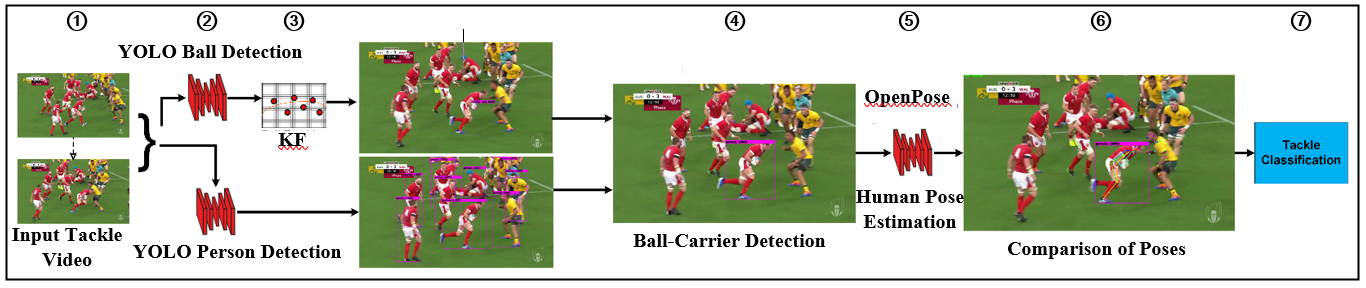}}\\
   \caption{Illustration representing the system overview. An input tackle video is passed through two YOLO networks (ball detection and person/player detection). The boundary box centre coordinates (x and y) are then filtered to track the ball - using a Kalman Filter (KF). The filter and person detection outputs are then used to determine the tackle frame, ball-carrier and tackler. The tackler and ball-carrier boundary box coordinates are then passed through the OpenPose network - producing key head points (head-centre). The head-centre of the ball-carrier and tackler are then compared and evaluated for the tackle type (high or low-risk).}
\label{fig:system_overview}
\end{figure*}

\section{Proposed Approach}
Automatic tackle risk evaluation during a rugby match has become a new challenge in the world of rugby. The aim of this research project was to explore and implement a computer vision approach to aid rugby referees in making objective decisions. Specifically, to influence their final ruling by classifying a tackle as either high or low-risk using the baseline criterion as mentioned above (focusing on head proximity). This paper is limited to focusing on the position of the player’s heads in determining the tackle risk. In this study, a high-risk tackle was defined as the orientation whereby the head-centre of the ball-carrier aligns with the head-centre of the tackler. Whereas a low-risk tackle is defined as the orientation whereby the head-centres do not align. 

The study focusses on implementing an unstudied approach by using markerless motion capture frameworks, namely: You Only Look Once version 4 (YOLO) Object Detection and Tensorflow OpenPose (OP) – along with in-game rugby union videos. A Kalman Filter (KF) was used as an intermediate step to refine detections made by the neural networks. The initial step was to determine the boundary box of the ball (using a YOLO trained network) and all persons/players within each frame. Using this information, the location of the ball-carrier, tackler and tackle frame could be determined. Thereafter, the boundary box coordinates were used in estimating the pose of the tackler and ball-carrier (using OpenPose). Finally, the pose estimates were used to evaluate the tackle – classifying it as either high or low-risk tackle (within the frame at which pre-contact is made). An overview of this system is shown in Figure \ref{fig:system_overview} and illustrated in the supplementary video.

\subsection{Data Collection}
The main source of data used to compile project models and test evaluations were qualitative. Therefore, the data comprised of images, video segments and a classification table provided by an analyst with expertise and experience in rugby video analysis. A detailed description of the data is described in the upcoming subsections. The project received its data from secondary data sources as the required images and videos were pre-existing (specifically in-game footage).

\subsubsection{Training Data}
A total of 551 images were used to train the ball detection network. These images were collected from open-source websites (royalty free images) such as OpenImage, Adobe Stock, MS-COCO dataset, Pexels, Pixabay and Unsplash. When exploring these databases, the search was refined using keywords: “sports”, “rugby”, and “ball”. Specifically, any image relating to the sport of rugby was selected as part of the training dataset (in-game, training, or any frolic relating to rugby). Therefore, any image containing a rugby ball - restricted to only include one rugby ball per image was included. In addition, these images were not restricted to any location, time, league, clothing type and colour, or player/person. Furthermore, the images were selected using various zoom levels varying from 36 to 1023 pixels in height and 16 to 1023 pixels in width – however, the entire shape of the ball was required in order to be included in the dataset. The camera angles of these images were not restricted while the image quality was limited to 1024x518 to 1024x1007. Furthermore, the various light intensity images were included  - limited to the ability for the user to see the rugby ball. These images along with annotations are provided as part of the paper\footnote{Link to ball detection data: \url{https://drive.google.com/drive/folders/1GvtlN-3CBxiI4CCvQW0dtIiupx39yEg_?usp=sharing}.}. 

\subsubsection{Testing/Input Data}
To evaluate the reliability of the system output, truth/ground data was required as a baseline of comparison. Multiple rugby union match videos (tackle segments) were provided to test the reliability of system’s output. These tackle segments were obtained from a registered rugby video analysis database and labelled (high or low-risk) by an analyst with expertise and experience in rugby video analysis \cite{hendricks2020consensus}.
A total of 109 tackle segments were provided which can be found at \textbf{link}.\footnote{Link to evaluated tackle segments: \url{https://drive.google.com/drive/folders/14VQ2s83sjGHFlXr4vfmQ_agU4XvF90R3?usp=sharing}.} These videos varied in quality (pixel dimensions 854 x 480 to 1920 x 1080), frames per second (25 to 30), zoom, angle (similar to the above mentioned). The length of these tackle segments begun from the point at which ball-carrier is visible to the point of pre-contact. The participants in these videos were limited to those part of the Rugby World Cup and Super Rugby teams. Furthermore, the video segments selected were front-on one-on-one tackles. It should be noted that the camera frame follows the ball-carrier (the camera is not stationary as it ensures that the ball-carrier is present in every frame). Tackle segments were omitted when the camera zoom was far out (players’ boundary box were $<$14\% of the video frame pixel height) and/or too much occlusion occurred. The ground truth data was not based on the referee’s outcome or ruling for the given tackle.

\subsection{You Only Look Once Object Detection (YOLO)}
The goal of YOLOv4 is to detect the location, confidence (accuracy of prediction model) and number of instances of an object in a single scan of an image \cite{16}. Therefore, it involves appending boundary boxes around the detected objects (categorising them in the corresponding object class) \cite{16}. The confidence score is calculated as shown in equation \eqref{eqn:(1)} which comprises of the intersection over union (of predicted and ground truth). Therefore, the confidence score is set to zero if there is no object detected. The probability of an object within a grid cell is defined as Pr (Object).

\begin{equation}
\label{eqn:(1)}
confidence = Pr(Object) \times \dfrac{area\_of\_overlap}{area\_of\_union}
\end{equation}

\subsubsection{Original Network}
The original YOLOv4 weights/network using the MS-COCO dataset was implemented on several videos. The Darknet-framework was used, with a network size of 608 × 608 (height x width) which produces a 65.7\% mAP@0.5 with 128.5 BFlops (while smaller networks such as 320 × 320 produced a 60.7\% mAP@0.5 with 35.5 BFlops). Therefore, detections were expected to be accurate and fast. This network was used to detect the location of all the humans/players within each frame (second step in Figure \ref{fig:system_overview}).

\subsubsection{Trained Network}
Images were labelled using LabelImg software by appending boundary boxes which allowed for YOLO annotation formatting (class\_index; x\_centre; y\_centre; width and height) parameters being measured in pixels. After labelling 551 images, the dataset was split 8:2 (train:test data).

Training parameters for ball detection (single class) were set using the recommendations provided by \cite{17}. Therefore, the learning rate was set to 0.001, batch size of 64, 64 subdivisions, network width and height to 352, iterations to 8000, and filter size (preceding the “yolo” layer in the neural network) was set to 18. Equations \eqref{eqn:(2)} and \eqref{eqn:(3)} represents the calculation used to determine the recommended (minimum) number of iterations and the filter size, respectively.

\begin{equation}
\label{eqn:(2)}
iterations= 2000 \times classes
\end{equation}

\begin{equation}
\label{eqn:(3)}
filter\_size =  (classes + 5) \times 3
\end{equation}

Methods of data augmentation were investigated to prevent the possibility of over-fitting the trained model. This was achieved by image adjustment of hue value of 0.1, tilt/rotation (10 degrees), saturation gradient (1.5), exposure (1.2), and mosaic (1). Normal distribution curves of the errors ($x$ and $y$) between the ground truth data and inferences made by the ball detection network were compiled. Errors were calculated using the centre coordinates of the boundary box ($x$ and $y$). A total of 547 points were compared. The resulting standard deviation for the x and y coordinate errors were 2.65 and 3.46, respectively. Furthermore, the resulting mean of the $x$ and $y$ coordinate errors were -0.43 and -0.09 pixels, respectively.

\subsection{Kalman Filter (KF)}
To optimally produce a boundary box estimate in the presence of noise and outliers, a KF \cite{13} was employed. For the sake of brevity we omit some details but the reader is encouraged to read \cite{forsyth2012computer} for further information on the KF for computer vision applications. 

A constant acceleration model was assumed with total of six state variables, generating the state matrix (x\_position; x\_velocity; x\_acceleration; y\_position; y\_velocity; y\_acceleration). It is recommended by \cite{20} that the square root of the process covariance matrix elements is the maximum change in corresponding acceleration. After investigating 22 video segments, a maximum change in the $x$ acceleration and $y$ acceleration was 6000 ${pixels/{s}^2}$ and 7000 ${pixels/{s}^2}$ respectively. The parameters for the measurement covariance matrix were obtained from the standard deviation of the YOLO coordinate error data. After fine-tuning adjustable parameters, it was observed that a standard deviation of 31 pixels in the $x$ and 15 pixels in the $y$, produced the best output.

When initializing the state matrix, the position states were set to the centre of the image (pixels), while the velocity and acceleration states were set to zero (both $x$ and $y$). The  initial values of the covariance matrix were 50 pixels for the position states, 10 pixels/s for the velocity states and the acceleration state covariance was set to 1 ${pixels/{s}^2}$ for both $x$ and $y$.

Outlier rejection was required to maintain the stability of the filter. A common approach used for outlier detection is to monitor the residual error (innovation) \cite{21}. This extends to evaluating whether the innovation remains within three standard deviations computed by the filter \cite{21}. To improve the number of reliable data measurements, the boundary was extended to five standard deviatons. The innovation \~{y}, covariance of the innovation S, and confidence score of the detection are required to evaluate whether the measured point is an outlier. Once the confidence score was less than 0.1 or/and the innovation for a measurement falls outside five times the square root of the innovation covariance, S, the innovation for that measurement was set to zero. Additionally, if the YOLO confidence score was less than $0.1$, the variance values of the measurement covariance matrix $R$ was set to the pixel dimensions and the maximum change in acceleration of both $x$ and $y$ would be set to 50 ${pixels/{s}^2}$. As real time estimation was not required, a Kalman Smoother was also implemented to further improve the system results \cite{20}.

\subsection{Ball-Carrier, Tackle Frame and Tackler Detection}
Using the filtered coordinates of the ball detection, the ball-carrier was determined by selecting the closest player to the ball. This was done by calculating the euclidean distance from each player’s boundary box centre to ball. The analyst indicated that a tackle frame is required to evaluate the risk of a tackle, this frame occurs within a seconds before the ball-carrier and tackler makes direct contact (pre-contact). The approach used in determining the tackle frame involved the use of boundary box overlap between “spectator” players and the ball-carrier.

Therefore, a counter registered the total number of overlaps throughout the tackle segment and the final overlap was labelled as the tackle frame. Furthermore, the tackler was identified as the player: (i) with the boundary box centre ($y$-coordinate) within the height range of the ball-carrier’s boundary box and  (ii) closest to the ball-carrier in the tackle frame (horizontal distance between boundary box centres). The resulting output is shown at step 4 in Figure \ref{fig:system_overview}.

\subsection{Pose Estimation And Tackle Evaluation}
Using the boundary box coordinates of the tackler and ball-carrier, the pose estimates for the two players were determined. These coordinates were passed through the Tensorflow OpenPose network where the pose estimates ($x$ and $y$ coordinates of 18 key points) of the tackler and ball-carrier were determined as shown in step 5 and 6 in Figure \ref{fig:system_overview}. OpenPose provides pose estimates of a person in a given frame, highlighting key points and generating a 2D skeleton \cite{22}. Given that high-risk tackles focus on the head location and orientation of the player/s, the pose estimates correspond $x$ and $y$ pixel coordinates of the left and right ear, left and right eye, and nose were averaged (head-centre).

It should be noted that OpenPose is subjected to potential errors when generating inferences (mislabelling key head points or not being able to generate a label when analysing a video). Therefore, in an attempt to reliably estimate the location of the ball-carrier's head-centre (in the tackle frame), the head-centres of the final three frames were averaged after being passed through a KF. The same filter parameters were used as the above mentioned KF with the exception of the measurement covariance matrix parameters (both $x$ and $y$ were set to 50 pixels), position state variance was set to 70 and evaluating the innovation within three standard deviations as less outliers were detected.

On the other hand, due to the limitation of not detecting the tackler in frames prior to the tackle frame - a filter could not be implemented. Therefore, to address the issue mentioned above, the y-coordinate of the head-centre produced by OpenPose was averaged with the estimated location of the tackler's head-centre. The estimation takes advantage that the head to body height ratio is 1:8 (12.5\% \~ 15\%) \cite{23}. Therefore, the pixel position of 7.5\% (assuming head-centre is the middle of the head) of the boundary box height (from the top) was used to obtain the y-coordinate head-centre estimate. This parameter was then averaged with the OpenPose head data.

\begin{figure}[t]
\fbox{\includegraphics[width=1\linewidth]{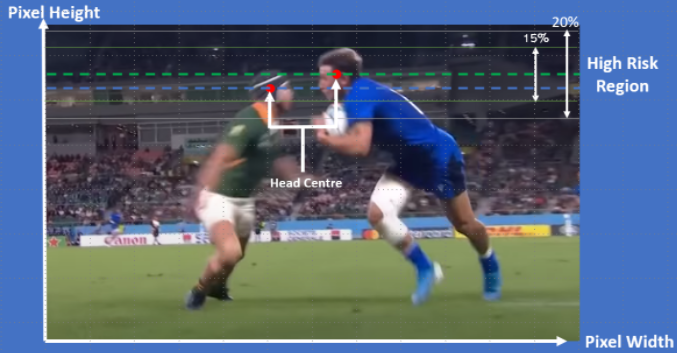}}\\
   \caption{Illustration of the tackle frame
   - highlighting the ball-carrier and tackler. Two high risk- regions are displayed (15\% and 20\%) with reference to the ball-carrier's head-centre. It should be noted that the head-centre of the tackler falls within both 15\% and 20\% high- risk region - classifying the tackle as a high-risk.}
\label{fig:risk_region}
\end{figure}

When evaluating the tackle, head-centres of the tackler and ball-carrier in the tackle frame (step 6 in Figure \ref{fig:system_overview}) are compared. If the head-centre (y-coordinate) of the tackler falls within the high-risk region (as shown in Figure \ref{fig:risk_region}), the tackle would be classified as a high-risk tackle (otherwise low-risk). The high risk-region is defined as boundary limit with the head-centre of the ball-carrier located in the middle. The head-centre of the ball-carrier was normalised to the height of the boundary box to the height of an average male. To obtain the best limits, the boundary limits were adjusted relative to the ball-carrier’s head-centre (5\%, 10\%, 15\%, 20\%, and 25\%). 

\section{Results and Discussion}
This section will evaluate the output of the KF and the reliability of the system output. In addition, the system limitation and shortcomings will be discussed and addressed.

\subsection{Kalman Filter Output}

\subsubsection{Ball Detection}
\begin{figure}[t]
\fbox{\includegraphics[width=1\linewidth]{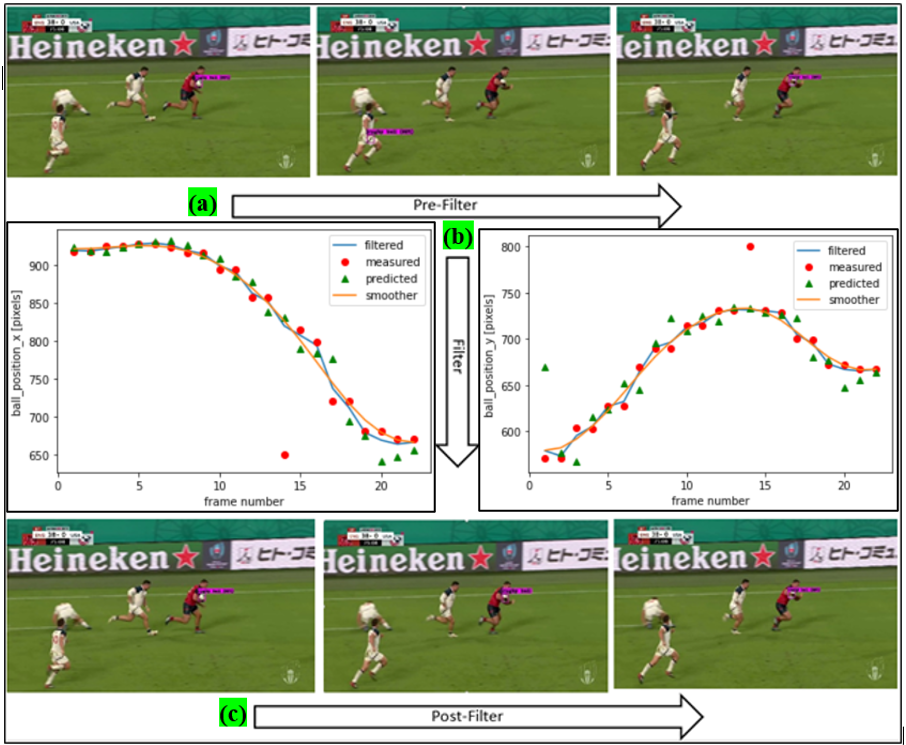}}\\
   \caption{Tackle segment displaying three frames (16, 17 and 18) of the YOLO ball detection network before and after applying the KF. The YOLO ball detection network mislabels frame 17 as an outlier whereby the KF rectifies.}
\label{fig:kf}
\end{figure}

Figure \ref{fig:kf} represents the input and output of the ball detection KF. The figure displays YOLO inferences made across three frames (16, 17 and 18) from a 24-frame tackle segment. When viewing the frames of the Pre-Filter in Figure \ref{fig:kf}a, it is observed that the YOLO network incorrectly identifies the  ball in the second image (frame 17). All YOLO  ball detection inferences made can be viewed in Figure \ref{fig:kf}b and can be seen that the measured coordinate in frame 17 is an outlier. Therefore, after filtering the coordinate inference points, it is observed that the algorithm correctly reduces the error between the measured and predicted data points. Furthermore, it is observed that the smoothing algorithm generates a refined transition dataset of the estimates from one frame to the next (Figure \ref{fig:kf}b). Finally, Figure \ref{fig:kf}c displays the final filtered output, Post-Filter (omitting and replacing the outlier with refined data point).

\subsubsection{Pose Estimation}
When implementing the KF for the pose estimates of the head-centres, some measured coordinates were outliers (in some instances the key points of the ball-carrier was an outlier on the tackle frame). Therefore, it was discovered that filtering and averaging (the corresponding coordinates of the head-centres in the final three frames) key points were required. The KF produced head-centre coordinates with improved accuracy in the tackle frame (corresponding to the ball-carrier’s head position).

\subsubsection{Challenges}
Although the KFs mentioned above refined the coordinates, many challenges and limitations reduced the robustness of the system. One key challenge  was the consecutive outliers detected within a dataset when implementing the KF for the ball detection network. This convinced the algorithm that the outlier points were part of the true dataset. A potential solution is to improve the ball detection training regime. This can be done by expanding the dataset such that more samples can be used for training and testing. Additionally, increasing the number of iterations with augmented data would also improve the inferences made by the model.\

Another challenge faced was the large magnitude of the change in acceleration parameters. Any large jump in the measured position would cause the filter to diverge and become unstable. One approach used to overcome this limitation was to adjust the change in acceleration relative to the confidence score produced by the YOLO network. Therefore, should the confidence score be less than 0.1 (indicating that the model is uncertain about detections), the change in acceleration would be set to 50 ${pixels/{s}^2}$. Another approach is to utilise techniques from a Moving Horizon Estimation as done in markerless 3D pose estimation \cite{joska2021Acino}.

\subsection{System Evaluation}
A total of 109 tackles were used and evaluated on the system. These tackles were selected based on the appropriate zoom level, camera angle and tackle type (head-on) of the recorded tackle. Of these 109 tackles, 58 of them ran successfully without any manual adjustments made to the system’s algorithm. Six additional tackles ran successfully after manually adjusting selected algorithm parameters. These parameters include adjusting the ball detection threshold (to 0.3), tackle frame (two/three frames before contact), and KF parameters (standard deviation in the x and y to 50 pixels) to obtain the correct tackler within the tackle frame. Therefore, a total of 64 tackles were successfully evaluated (73.4\% classified as low-risk tackles by the analyst). 

The values in Table 1 represent the overall tackle accuracy calculated using equation \eqref{eqn:(11)} . The accuracy calculation was done using the total number of tackles (109) and the total number of tackles that ran successfully without any adjustments (64). Furthermore, the accuracy calculation was done for multiple high-risk regions (5\%, 10\%, 15\%, 20\% and 25\%) containing high and low-risk tackles.

\begin{equation}
\label{eqn:(11)}
{Accuracy\_of\_Tackles} = \dfrac{{Correct\_Detections}}{{Total\_Number\_of\_Tackles}}
\end{equation}

\begin{table}
\begin{center}
\begin{tabular}{ |p{1.5cm}||p{0.9cm}|p{0.9cm}|p{0.9cm}|p{0.9cm}|p{0.9cm}|}
\hline
 & 5\% & 10\% & 15\% & 20\% & 25\% \\
\hline\hline
\textbf{Accuracy (n=109)} & 39.44\% & 37.61\% & 36.70\% & 33.03\% & 30.28\% \\
\hline
\textbf{Accuracy (n=64)} & 67.19\% & 64.06\% & 62.50\% & 56.25\% & 51.56\% \\
\hline
\textbf{Recall)} & 29.41\% & 47.06\% & 70.59\% & 70.59\% & 82.35\% \\
\hline
\textbf{Precision} & 35.71\% & 36.36\% & 38.71\% & 34.29\% & 33.33\% \\
\hline
\textbf{F1 Score} & 0.32 & 0.41 & 0.50 & 0.46 & 0.47 \\
\hline
\end{tabular}
\end{center}
\caption{Evaluation of overall tackle accuracy, recall, precision and F1 Score for high-risk regions of 5\%, 10\%, 15\%, 20\% and 25\% using the total number of tackles (109) and evaluated tackles (64).}
\end{table}

It is observed that the high-risk region of 5\% produces the highest overall tackle accuracy in both total and evaluated tackles. A possible data influence is that 26.6\% of the total tackles are high-risk (17/64). Therefore, the 5\% region is most likely to classify a tackle as low-risk given it is the smallest high-risk region compared to the 10\% to 25\% regions (small window for both tackler and ball-carrier head-centres to be located). However, it is observed that the 25\% high-risk region obtains the highest high-risk accuracy (82.35\%) while the 5\% high-risk region obtains the highest low-risk accuracy (80.85\%). Therefore, the 15\% high-risk region represents a midway point for high-risk and low-risk tackle accuracy. 

Given that the low-risk tackles greatly outnumber the number of high-risk tackles, it can be viewed as an imbalanced classification problem. Therefore, the recall statistic would model the relevant cases (high-risk tackles) in the dataset. To compensate for the low precision of the recall metric, the actual relevant data (influenced by the proportion of points) should be determined (precision). Therefore, the F1 score finds the harmonic mean (optimal balance) between precision and recall. With CHD, HLL and LLH being the correct high-risk detections, the high-risk tackles labelled as low-risk and the low-risk tackles labelled as high-risk, respectively - the values of recall, precision and F1 Score (equation \eqref{eqn:(12)}) are shown in Table 1.

\begin{equation}
\label{eqn:(12)}
{F1} = \dfrac{{CHD}}{{ CHD + 0.5\times (HLL+LLH)}}
\end{equation}

Given that the F1 Score is determined with an equal contribution in precision and recall when evaluating the system’s performance – the higher the score, the better the performance. Therefore, it is observed from Table 1 that the high-risk region of 15\% has the highest F1 Score (0.5) compared to the other high-risk regions. Furthermore, represent the reliability of the system accurately – the Cohen’s Kappa coefficient (CKC) is determined.

Table 2 and 3 represents the CKC for all five high-risk regions using the 64 successfully evaluated tackle. The CKC matrix (Table 2) compares the classification of the system output to the ground truth data. Each cell contains an array of five numbers, these represent the number classifications made for the high-risk region of [5\%, 10\%, 15\%, 20\%, 25\%]. Furthermore, Table 3 represents probability parameters (probability of agreement [P\_0], probability of random agreement [P\_e], probability of high-risk tackles [P\_HR] and probability low-risk tackles  [P\_LR])used to determine the inter-rater reliability coefficient, Cohen’s Kappa for all high-risk regions.

\begin{table}
\begin{center}
\begin{tabular}{ |p{1.5cm}||p{1.5cm}|p{1.5cm}|p{1.5cm}|}
 \hline
 \multicolumn{4}{|c|}{\textbf{Ground Truth}} \\
 \hline
&  &High Risk&Low Risk\\
 \hline
\textbf{System Output}& High Risk & [5, 8, 12, 12, 14]&[9 ,14, 19, 23, 28]\\
\hline
& Low Risk & [12, 9, 5, 5, 3]&[38, 33, 28, 24, 19]\\
 \hline
\end{tabular}
\end{center}
\caption{CKC confusion matrix representing the classification of the system output compared to the ground truth data (high-risk region indices of array corresponds to: [5\%, 10\%, 15\%, 20\%, 25\%]).}
\end{table}

\begin{table}
\begin{center}
\begin{tabular}{|p{1.3cm}||p{0.9cm}|p{0.9cm}|p{0.9cm}|p{0.9cm}|p{0.9cm}|}
\hline
Parameter&
Value (HRR 5\%)&
Value (HRR 10\%)&
Value (HRR 15\%)&
Value (HRR 20\%)&
Value (HRR 25\%)\\
\hline\hline
P\_0&
0.67&
0.66&
0.64&
0.57&
0.53\\
P\_{HR}&
0.06&0.10&0.14&0.16&0.19\\
P\_{LR}&
0.56&
0.47&
0.36&
0.31&
0.24\\
P\_e&
0.62&
0.57&
0.50&
0.47&
0.42\\
CKC&
0.13&
0.20&
0.28&
0.19&
0.19\\
\hline
\end{tabular}
\end{center}
\caption{Table highlighting the probability parameters which are used to determine the inter-rater reliability coefficient, Cohen’s Kappa. All high-risk region (HRR) values are displayed (5\%, 10\%, 15\%, 20\%, 25\%).}
\end{table}

It is observed in both instances, the high-risk region of 15\% produces the highest Cohen’s Kappa coefficient indicating the highest level of agreement between the system output and the ground truth data. Therefore a CKC of 0.28 indicates a fair level of reliability while the other high risk-regions indicates a non to slight level of reliability.

\begin{figure}[t]
\fbox{\includegraphics[width=1\linewidth]{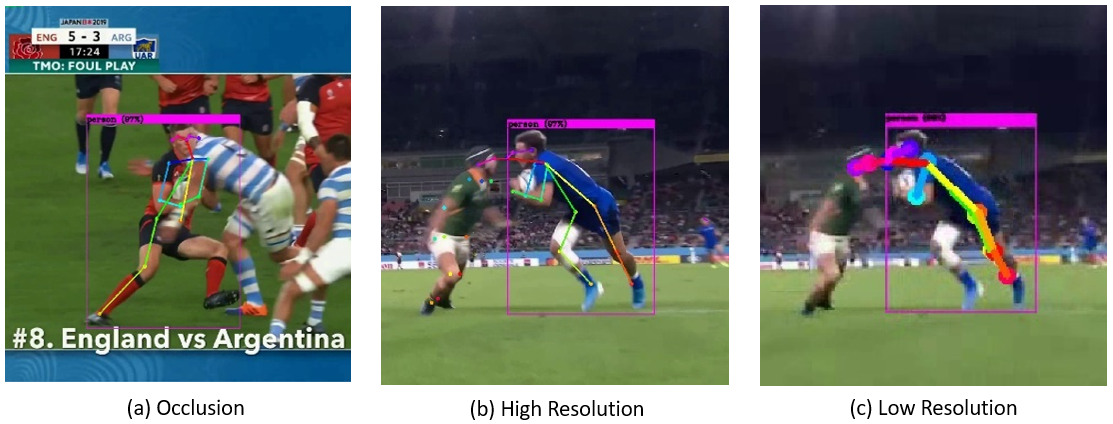}}\\
   \caption{Illustration representing various challenges experienced by the model. Occlusion (a) is observed as the ball-carrier and tackler share the same skeleton generated by OpenPose. The influence on video quality on the inferences made is shown in (b) and (c) - higher resolution indicates an improved key point mapping.}
\label{fig:challenges}
\end{figure}

\subsubsection{Challenges}
There were many reasons for the system only being able to evaluate 64 out of the 109 tackles. Most reasons were closely linked to the shortcomings of  YOLO and OpenPose (inferences). One of the biggest limitations was the occlusion surrounding the ball-carrier or tackler within the tackle frame. This made it difficult for YOLO and OpenPose to identify individual players and generate separate sets of key points (when two players overlap - OpenPose tends to generate key points for one player across two people as shown in Figure \ref{fig:challenges}a), respectively. This issue limited the frame at which the tackle occurs - occlusion will occur if the players are too close. Furthermore, the problem extends to errors in identifying the tackle frame as the pose estimates overlap between ball-carrier and tackler before their boundary boxes overlap (suggesting that frames before the tackle frame should be used). In addition, the task of correctly identifying the ball and ball-carrier in every frame becomes challenging. These challenges were addressed by using a KF as mentioned above.

Video quality stemmed as an additional obstacle for the YOLO and OpenPose network when generating inferences. The reason for this is that the network is not familiar with low quality images (OpenPose), making it much harder for the network to identify key points. This is seen in Figure \ref{fig:challenges}b and \ref{fig:challenges}c whereby the OpenPose network identifies more key points in the high resolution image compared to the low resolution one. This issue can be addressed by re-training the OpenPose network (from its latest weights) using low quality images.

\section{Future Work}
Occlusion is one of the biggest challenges with the proposed approach. This problem is important to address as rugby tackle footage always contain occlusions. A potential solution to this issue is to explore alternative multi-person pose estimation frameworks (with improved detections when players overlap or come into contact). Furthermore, a new model could be trained which identifies the pose of players before and during a rugby tackle (a classifier model detecting the tackle frame could also be investigated). Furthermore, in an attempt to improve the tackler detection, efforts could focus on determining which player's boundary box overlaps the most with tackler in the previous frame (and back track the location of the tackler using the same procedure). To improve the quality of the model, a larger dataset should be used - specifically increasing the number of high-risk tackles. In addition, efforts could be focused on not only assisting the referee's decision but predict the final ruling (legal or illegal) by using the entire pose of the tackler and ball-carrier. 

\section{Conclusion}
With the increasing frequency of concussion incidents in the world of rugby, various protocols are being investigated and implemented to prevent future concussions. Most protocols focus on improving the orientation of the tackler when engaging in a tackle. This paper suggests an automatic tackle risk classification approach using YOLO and OpenPose. This method aims at assisting referees in making objective decisions of foul play. The system does this by estimating the pose of the tackler and ball-carrier, thereafter, evaluating the pose of their head centres. The system will either indicate a tackle is high-risk or low-risk which is influenced by the high-risk region. When analysing the results, the 15\% high-risk region was identified as the most reliable due to finding a midway between high-risk and low-risk tackle accuracy, has the highest F1 Score (0.50), and the highest CKC (0.28) – it was used as the final output to the system.

Many challenges became apparent from the system dynamics including: presence of occlusion, low quality videos, multiple outliers, and fine tuning the KF parameters to maintain its stability. Therefore, further research may be conducted to address the classification system shortcomings. This paper serves as a proof-of-concept for the implementation of markerless motion capture techniques within the field of collision-based sports. An example of this is to generate a neural network which directly classifies a tackle (high or low-risk) possibly using a recurrent neural network \cite{schuster1997bidirectional}.

{\small
\bibliographystyle{ieee}
\bibliography{egbib}

\begin{thebibliography}{10}\itemsep=-1pt

\bibitem{17}
AlexeyAB.
\newblock Yolo v4, v3 and v2 for windows and linux.
\newblock \url{https://github.com/AlexeyAB/darknet}, 2020.
\newblock Accessed: 2021-03-11.

\bibitem{brooks2005epidemiology}
J.~H. Brooks, C.~Fuller, S.~Kemp, and D.~B. Reddin.
\newblock Epidemiology of injuries in english professional rugby union: part 1
  match injuries.
\newblock {\em British journal of sports medicine}, 39(10):757--766, 2005.

\bibitem{burger2020lay}
N.~Burger, M.~Lambert, and S.~Hendricks.
\newblock Lay of the land: narrative synthesis of tackle research in rugby
  union and rugby sevens.
\newblock {\em BMJ Open Sport \& Exercise Medicine}, 6(1):e000645, 2020.

\bibitem{22}
Z.~Cao, G.~Hidalgo, T.~Simon, S.-E. Wei, and Y.~Sheikh.
\newblock Openpose: realtime multi-person 2d pose estimation using part
  affinity fields.
\newblock {\em IEEE transactions on pattern analysis and machine intelligence},
  43(1):172--186, 2019.

\bibitem{forsyth2012computer}
D.~A. Forsyth and J.~Ponce.
\newblock {\em Computer vision: a modern approach}.
\newblock Pearson,, 2012.

\bibitem{fuller2010injury}
C.~W. Fuller, T.~Ashton, J.~H. Brooks, R.~J. Cancea, J.~Hall, and S.~P. Kemp.
\newblock Injury risks associated with tackling in rugby union.
\newblock {\em British journal of sports medicine}, 44(3):159--167, 2010.

\bibitem{fuller2015epidemiology}
C.~W. Fuller, A.~Taylor, and M.~Raftery.
\newblock Epidemiology of concussion in men's elite rugby-7s (sevens world
  series) and rugby-15s (rugby world cup, junior world championship and rugby
  trophy, pacific nations cup and english premiership).
\newblock {\em British journal of sports medicine}, 49(7):478--483, 2015.

\bibitem{fuller2017accuracy}
G.~Fuller, S.~Kemp, and M.~Raftery.
\newblock The accuracy and reproducibility of video assessment in the
  pitch-side management of concussion in elite rugby.
\newblock {\em Journal of science and medicine in sport}, 20(3):246--249, 2017.

\bibitem{gardner2015systematic}
A.~Gardner, G.~L. Iverson, C.~R. Levi, P.~W. Schofield, F.~Kay-Lambkin, R.~M.
  Kohler, and P.~Stanwell.
\newblock A systematic review of concussion in rugby league.
\newblock {\em British journal of sports medicine}, 49(8):495--498, 2015.

\bibitem{gardner2014systematic}
A.~J. Gardner, G.~L. Iverson, W.~H. Williams, S.~Baker, and P.~Stanwell.
\newblock A systematic review and meta-analysis of concussion in rugby union.
\newblock {\em Sports medicine}, 44(12):1717--1731, 2014.

\bibitem{10}
A.~J. Gardner, R.~Kohler, W.~McDonald, G.~W. Fuller, R.~Tucker, and
  M.~Makdissi.
\newblock The use of sideline video review to facilitate management decisions
  following head trauma in super rugby.
\newblock {\em Sports medicine-open}, 4(1):1--8, 2018.

\bibitem{gardner2018use}
A.~J. Gardner, R.~Kohler, W.~McDonald, G.~W. Fuller, R.~Tucker, and
  M.~Makdissi.
\newblock The use of sideline video review to facilitate management decisions
  following head trauma in super rugby.
\newblock {\em Sports medicine-open}, 4(1):1--8, 2018.

\bibitem{hendricks_2020}
S.~Hendricks, Nov 2020.

\bibitem{hendricks2021tackle}
S.~Hendricks, B.~Jones, and N.~Burger.
\newblock Tackle injury epidemiology and performance in rugby league--narrative
  synthesis.
\newblock {\em South African Journal of Sports Medicine}, 33(1):1--8, 2021.

\bibitem{hendricks2020consensus}
S.~Hendricks, K.~Till, S.~Den~Hollander, T.~N. Savage, S.~P. Roberts,
  G.~Tierney, N.~Burger, H.~Kerr, S.~Kemp, M.~Cross, et~al.
\newblock Consensus on a video analysis framework of descriptors and
  definitions by the rugby union video analysis consensus group.
\newblock {\em British journal of sports medicine}, 54(10):566--572, 2020.

\bibitem{joska2021Acino}
D.~Joska, L.~Clark, N.~Muramatsu, R.~Jericevich, F.~Nicolls, A.~Mathis, M.~W.
  Mathis, and A.~Patel.
\newblock Acinoset: A 3d pose estimation dataset and baseline models for
  cheetahs in the wild.
\newblock {\em arXiv preprint arXiv:2103.13282}, 2021.

\bibitem{king2010video}
D.~King, A.~P. Hume, and T.~Clark.
\newblock Video analysis of tackles in professional rugby league matches by
  player position, tackle height and tackle location.
\newblock {\em International Journal of Performance Analysis in Sport},
  10(3):241--254, 2010.

\bibitem{20}
R.~Labbe.
\newblock Kalman and bayesian filters in python.
\newblock {\em Chap}, 7:246, 2014.

\bibitem{23}
G.~Mather.
\newblock Head--body ratio as a visual cue for stature in people and sculptural
  art.
\newblock {\em Perception}, 39(10):1390--1395, 2010.

\bibitem{12}
T.~B. Moeslund, G.~Thomas, and A.~Hilton.
\newblock {\em Computer vision in sports}, volume~2.
\newblock Springer, 2014.

\bibitem{14}
H.~Nguyen, F.~Ayachi, E.~Goubault, C.~Lavigne-Pelletier, B.~J. McFadyen, and
  C.~Duval.
\newblock Longitudinal study on the detection and evaluation of onset mild
  traumatic brain injury during dual motor and cognitive tasks.
\newblock In {\em icSPORTS}, pages 77--83, 2015.

\bibitem{quarrie2008tackle}
K.~L. Quarrie and W.~G. Hopkins.
\newblock Tackle injuries in professional rugby union.
\newblock {\em The American journal of sports medicine}, 36(9):1705--1716,
  2008.

\bibitem{raftery2020getting}
M.~Raftery, R.~Tucker, and {\'E}.~C. Falvey.
\newblock Getting tough on concussion: how welfare-driven law change may
  improve player safety—a rugby union experience, 2020.

\bibitem{16}
J.~Redmon, S.~Divvala, R.~Girshick, and A.~Farhadi.
\newblock You only look once: Unified, real-time object detection.
\newblock In {\em Proceedings of the IEEE conference on computer vision and
  pattern recognition}, pages 779--788, 2016.

\bibitem{11}
W.~Rugby.
\newblock Television match official (tmo) global trial protocol ..., Aug 2019.

\bibitem{schuster1997bidirectional}
M.~Schuster and K.~K. Paliwal.
\newblock Bidirectional recurrent neural networks.
\newblock {\em IEEE transactions on Signal Processing}, 45(11):2673--2681,
  1997.

\bibitem{spitz2021video}
J.~Spitz, J.~Wagemans, D.~Memmert, A.~M. Williams, and W.~F. Helsen.
\newblock Video assistant referees (var): The impact of technology on decision
  making in association football referees.
\newblock {\em Journal of Sports Sciences}, 39(2):147--153, 2021.

\bibitem{stokes2021does}
K.~A. Stokes, D.~Locke, S.~Roberts, L.~Henderson, R.~Tucker, D.~Ryan, and
  S.~Kemp.
\newblock Does reducing the height of the tackle through law change in elite
  men’s rugby union (the championship, england) reduce the incidence of
  concussion? a controlled study in 126 games.
\newblock {\em British journal of sports medicine}, 55(4):220--225, 2021.

\bibitem{15}
Y.~Tanabe, T.~Kawasaki, H.~Tanaka, K.~Murakami, K.~Nobuhara, T.~Okuwaki, and
  K.~Kaneko.
\newblock The kinematics of 1-on-1 rugby tackling: a study using 3-dimensional
  motion analysis.
\newblock {\em Journal of shoulder and elbow surgery}, 28(1):149--157, 2019.

\bibitem{21}
J.-A. Ting, E.~Theodorou, and S.~Schaal.
\newblock A kalman filter for robust outlier detection.
\newblock In {\em 2007 IEEE/RSJ International Conference on Intelligent Robots
  and Systems}, pages 1514--1519. IEEE, 2007.

\bibitem{tucker2017risk}
R.~Tucker, M.~Raftery, S.~Kemp, J.~Brown, G.~Fuller, B.~Hester, M.~Cross, and
  K.~Quarrie.
\newblock Risk factors for head injury events in professional rugby union: a
  video analysis of 464 head injury events to inform proposed injury prevention
  strategies.
\newblock {\em British journal of sports medicine}, 51(15):1152--1157, 2017.

\bibitem{13}
W.~Wang, N.~He, K.~Yao, and J.~Tong.
\newblock Improved kalman filter and its application in initial alignment.
\newblock {\em Optik}, 226:165747, 2021.

\bibitem{west2020trends}
S.~W. West, L.~Starling, S.~Kemp, S.~Williams, M.~Cross, A.~Taylor, J.~H.
  Brooks, and K.~A. Stokes.
\newblock Trends in match injury risk in professional male rugby union: a
  16-season review of 10 851 match injuries in the english premiership
  (2002--2019): the rofessional ugby njury urveillance roject.
\newblock {\em British journal of sports medicine}, 2020.

\bibitem{williams2013meta}
S.~Williams, G.~Trewartha, S.~Kemp, and K.~Stokes.
\newblock A meta-analysis of injuries in senior men’s professional rugby
  union.
\newblock {\em Sports medicine}, 43(10):1043--1055, 2013.

\end{thebibliography}
}

\end{document}